\pdfoutput=1
\documentclass[11pt]{article}

\usepackage[final]{acl}

\usepackage{times}
\usepackage{latexsym}
\usepackage{float}
\usepackage{tabularx} 
\usepackage[T1]{fontenc}
\usepackage[utf8]{inputenc}

\usepackage{microtype}

\usepackage{inconsolata}

\usepackage{booktabs}
\usepackage{graphicx}
\usepackage{longtable}
\usepackage{tcolorbox}
\usepackage{stackengine}
\usepackage{multirow}
\usepackage{graphicx}
\usepackage{wrapfig}
\usepackage{hyperref}
\usepackage{makecell}
\usepackage{booktabs} % For better horizontal lines
\usepackage{array}    % For specifying widths
\usepackage{geometry} % Optional: to adjust page margins if needed

\usepackage{url}            % simple URL typesetting
\usepackage{amsfonts}       % blackboard math symbols
\usepackage{nicefrac} 
\usepackage[figure]{hypcap}
\usepackage{microtype}      % microtypography
\usepackage{xcolor,colortbl}         % colors
\usepackage{times}
\usepackage{latexsym}
\usepackage{multicol}
\usepackage{blindtext}
\usepackage{tabu}
\usepackage{amsmath, bm}

\usepackage{subcaption}
\usepackage{caption}
\usepackage[normalem]{ulem}
\usepackage{soul}
\usepackage[shortlabels]{enumitem}
\usepackage{array}
\usepackage{pgffor}
\usepackage{textcomp}
\usepackage{amssymb}
\usepackage{pifont}
\usepackage{booktabs}
\usepackage{tabularx}

\definecolor{lightgreen}{rgb}{0.8, 0.95, 0.8}
\definecolor{lightred}{rgb}{0.95, 0.8, 0.8}
\definecolor{naplesyellow}{rgb}{0.98, 0.85, 0.37}
\definecolor{pastelyellow}{rgb}{0.99, 0.99, 0.59}

\title{Parsing the Switch: LLM-Based UD Annotation for Complex Code-Switched and Low-Resource Languages}

\author{
  Olga Kellert$^{1}$\thanks{Equal contribution.} \quad
  Nemika Tyagi$^{1}$\footnotemark[1] \quad
  Muhammad Imran$^2$
  \textbf{Nelvin Licona-Guevara}$^1$ \quad \\
  \textbf{Carlos Gómez-Rodríguez}$^2$ \\
  $^1$Arizona State University \quad $^2$Universidade da Coruña, CITIC \\
  \small{\texttt{\{olga.kellert, ntyagi8, nliconag\}@asu.edu}}, \small{\texttt{\{m.imran, carlos.gomez\}@udc.es}}
}

\begin{document}
\maketitle
\begin{abstract}

Code-switching presents a complex challenge for syntactic analysis, especially in low-resource language settings where annotated data is scarce. While recent work has explored the use of large language models (LLMs) for sequence-level tagging, few approaches systematically investigate how well these models capture syntactic structure in code-switched contexts. Moreover, existing parsers trained on monolingual treebanks often fail to generalize to multilingual and mixed-language input. To address this gap, we introduce the \textit{BiLingua Parser}, an LLM-based annotation pipeline designed to produce Universal Dependencies (UD) annotations for code-switched text. First, we develop a prompt-based framework for Spanish-English and Spanish-Guaraní data, combining few-shot LLM prompting with expert review. Second, we release two annotated datasets, including the first Spanish-Guaraní UD-parsed corpus. Third, we conduct a detailed syntactic analysis of switch points across language pairs and communicative contexts. 
%including <add some observation>. 
Experimental results show that BiLingua Parser achieves up to \textbf{95.29\%} LAS after expert revision, significantly outperforming prior baselines and multilingual parsers. These results show that LLMs, when carefully guided, can serve as practical tools for bootstrapping syntactic resources in under-resourced, code-switched environments\footnote{Data and source code are available at \url{https://github.com/N3mika/ParsingProject}.}.

%Code-switching, the practice of alternating between two or more languages within a single utterance or discourse, presents significant challenges for natural language processing tasks such as dependency parsing. This paper has three main goals: 1) introducing LLM-based linguistic annotation for analyzing syntactic structure in Spanish-Guarani and Spanish-English code-switched communication, 2) realising two datasets with PoS tags and Universal Dependencies annotated on code-switched communication 3) demonstrating the utility of these datasets by leveraging dependency parsing for a syntactic analysis of code-switch points. We developed and iteratively refined a few-shot prompting strategy to generate high-quality dependency annotations, achieving strong agreement with expert judgments. To assess the performance of our method, we also trained a Universal Dependencies (UD) parser based on Sequence Labeling (UDSL) on available monolingual and synthetic code-switched data. While the UDSL parser struggled with syntactic structures at code-switch points, our LLM-based annotation is consistently more accurate in these regions. These findings highlight the potential of large language models to support code-switching-aware parsing, especially for under-resourced language pairs.

\end{abstract}

\section{Introduction}
\label{sec:intro}
Code-switching (CSW) is a widespread linguistic phenomenon observed in multilingual communities around the world. Despite its prevalence in spoken and informal digital communication, it remains a complex challenge for natural language processing (NLP), particularly for syntactic parsing. One of the central issues is that most state-of-the-art parsing models are trained on monolingual treebanks and thus lack robustness when applied to mixed-language data~\citep{ozates-etal-2022-improving}.

\noindent Previous works of \citet{ozates-etal-2022-improving, rijhwani2017analyzing, bhat-etal-2018-universal}  took an important step toward addressing this gap by proposing, for instance, a semi-supervised dependency parsing framework that augments training with auxiliary sequence labeling tasks \citep{ozates-etal-2022-improving}. Their model improved parsing accuracy on Turkish-German spoken corpus by learning better representations of syntactic structure in a multilingual setting. However, even with such enhancements, existing models often rely on large amounts of annotated data, which is particularly limiting for under-resourced language pairs.

\begin{figure}[t!]
    \centering
    \includegraphics[width=7.5cm]{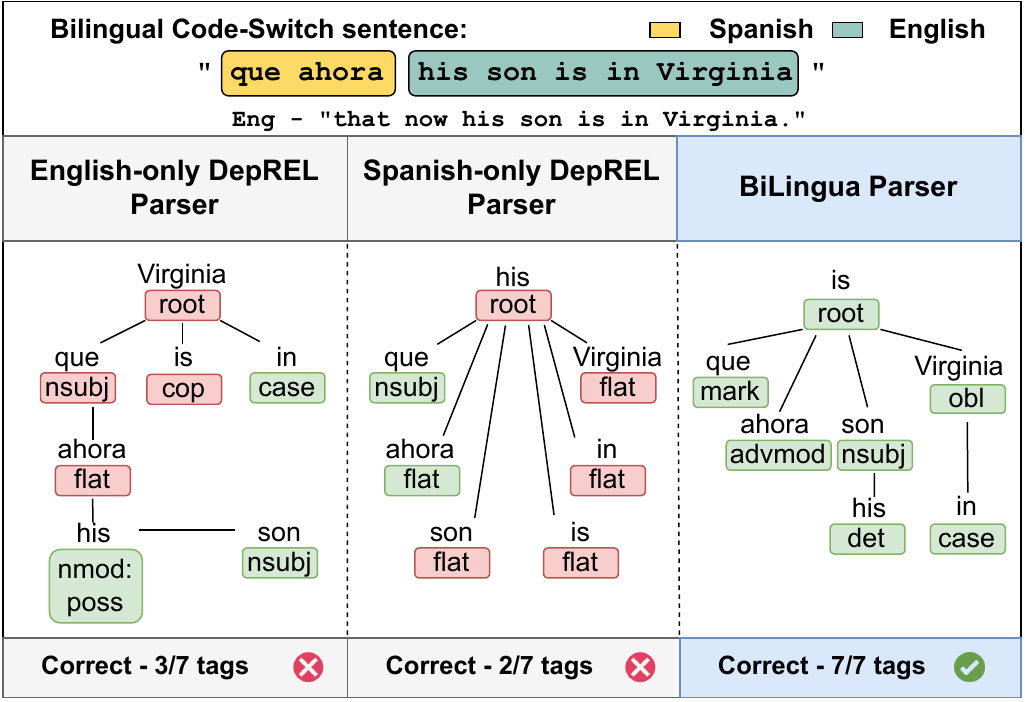}
   \caption{Comparison of dependency relation predictions (DepREL) for a Spanish-English CSW sentence across three parsers. The English-only and Spanish-only models misassign key relations due to monolingual bias. In contrast, the \textit{BiLingua Parser} correctly analyzes the full structure across the language boundary.}
    \label{fig:fig1}
\end{figure}

\begin{figure*}
    \centering
    \includegraphics[width=1.0\linewidth]{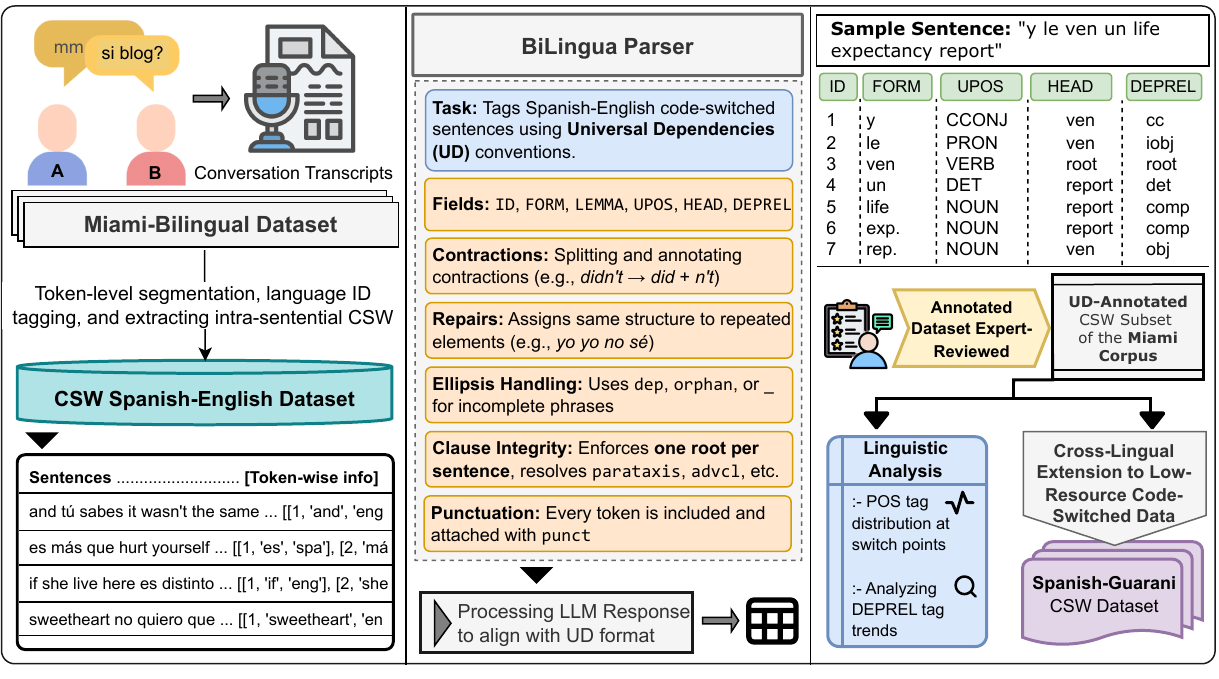}
 \caption{Overview of BiLingua Parser pipeline for Spanish-English code-switching. \textbf{Left:} Conversation transcripts from the Miami-Bilingual Corpus are processed through token-level segmentation, language ID tagging, and filtering of intra-sentential code-switches. \textbf{Center:} The \textit{BiLingua Parser} assigns UD tags to CSW sentences, handling contractions, repetitions, ellipsis, and clausal structure. \textbf{Right:} The resulting annotated dataset is reviewed by linguistic experts and enables downstream tasks such as POS/DEPREL analysis and extension to low-resource settings, including Spanish-Guaraní.}
    \label{fig:teaser}
\end{figure*}

Motivated by this lack of resources, we introduce BiLingua Parser, a bilingual syntactic parser based on large language models (LLMs), specifically the GPT-4.1 model, to generate syntactically annotated CSW datasets. Figure \ref{fig:fig1} illustrates how current monolingual parsers perform significantly worse than BiLingua Parser on a widely studied CSW language pair. Next, we tackle this issue for two language pairs in particular: Spanish-English (a relatively well-resourced code-switching language pair) and Spanish-Guarani (a low-resource language pair for which most linguistic tools are largely unavailable) \cite{chiruzzo2023overviewguaspaiberlef2023}.  To our knowledge, these are the first datasets for Spanish-English and Spanish–Guaraní code-switching with UD-based syntactic annotations reviewed by native speakers. The entire pipeline for creating and using BiLingua Parser is shown in Figure \ref{fig:teaser}.

%We demonstrate that LLM-based annotation, prompted with linguistic rules and a few gold-annotated examples, are capable of producing high-quality part-of-speech and dependency labels \nemika{head, and head ids} even in challenging code-switching environments.
In addition to developing BiLingua parser, we also examine the limitations of current 
%evaluation metrics used for linguistic parsers. 
syntactic parsing evaluation metrics.
%To address this, 
To this end,
we introduce additional methods for assessing the performance of our parser with the help of linguistic experts. Moreover, the annotated datasets we create also support a wide range of linguistic analyses involving bilingual speakers, including understanding of fine-grained switch-point behavior. While most previous structural studies in NLP on code-switching have focused on part-of-speech (POS) tags \cite{soto2020codeswitching, rijhwani2016estimating, solorio2008pos}, our analysis using dependency parsing shows that syntactic subjects (nsubj) are among the most frequent switch points in both language pairs and that Spanish-Guarani code-switching 
%offers a 
exhibits
higher variation in switch points. This finding underscores the value of dependency parsing for analyzing switch points across languages.
The main contributions of our work are as follows:
\begin{itemize}
    \item We introduce a method for generating UD-style syntactic annotations using LLM-based prompting, and compare it against baselines from previous work on dependency parsing of code-switched texts, as well as a parser trained on a synthetic combination of monolingual treebanks.
    \item We release two code-switched datasets, with POS and dependency annotations, reviewed by native speakers, including a new resource for Spanish- Guaraní.
    \item We conduct a linguistic case study of common syntactic structures at code-switch boundaries, revealing cross-linguistic switching patterns.
\end{itemize}
Overall, we believe our findings and released resources will support future work in both computational modeling and linguistic analysis of code-switching, especially for under-resourced and typologically diverse language pairs.
 %To evaluate our approach, we compare the LLM output to the results of a trained Universal Dependencies parser based on Sequence Labeling (UDSL) applied to the same data. Our findings show that the LLM-based system achieves far superior performance at code-switch points and in low-resource segments than UDSL and previous systems based on code-switched treebanks, providing a viable pathway for enhancing multilingual parsing without the need for extensive treebanks. \nemika{-needs to be rephrased}

\section{Related Work}
\label{sec:related}

Parsing code-switched text is considerably more challenging than monolingual parsing due to structural variability, mixed grammar rules, and limited annotated corpora. Prior work~\citep{rijhwani2017analyzing, solorio2008pos, solorio2014overview, ozates-etal-2022-improving} has demonstrated that models trained exclusively on monolingual data perform poorly on CSW without specific adaptation. \citet{ozates-etal-2022-improving} addressed this gap by proposing a semi-supervised parsing framework that incorporates auxiliary sequence labeling tasks. Their approach improved parsing performance on Turkish-German CSW with labeled attachment scores (LAS) reaching up to 73\%. Similarly, \citet{bhat-etal-2018-universal} and \citet{rijhwani2017analyzing} reported LAS scores in the 70–72\% range for Hindi-English data using adapted models (Table~\ref{tab:cs_baseline_results}).

\begin{table}[tbp]
\centering
\resizebox{\columnwidth}{!}{
\begin{tabular}{llll}
\toprule
\textbf{Study} & \textbf{Language Pair} & \textbf{POS Accuracy} & \textbf{LAS (Parsing)} \\
\midrule
\citet{solorio2014overview} & Spanish-English & 85.1\% & -- \\
\citet{solorio2014overview} & Hindi-English   & 83.3\% & -- \\
\citet{rijhwani2017analyzing} & Spanish-English & 80.0\% & -- \\
\citet{ozates-etal-2022-improving} & Hindi-English & -- & 71.93\% \\
\citet{ozates-etal-2022-improving} & Turkish-German & -- & 73.0\% \\
\citet{bhat-etal-2018-universal} & Hindi-English & -- & 71.03\% \\
\bottomrule
\end{tabular}
}
\caption{POS tagging and dependency parsing performance on code-switched datasets in prior work.}
\label{tab:cs_baseline_results}
\end{table}

Most prior research focuses on part-of-speech tagging rather than full syntactic parsing, and available resources remain limited to a few language pairs. A draft UD treebank exists for Spanish-English code-switching, but it is not publicly released, further illustrating the scarcity of syntactically annotated CSW data. Efforts to increase parsing speed, such as recasting dependency parsing as a sequence labeling task~\citep{Strzyz2019ViableDP, Roca2023ASF}, have improved runtime, but still depend heavily on monolingual training data.

Our work builds on these foundations but shifts toward LLM-based annotation. Large language models like OpenAI GPT can be prompted with examples and linguistic rules to produce annotations without requiring extensive supervised training. We apply this technique to generate and evaluate new CSW datasets, including one for Spanish-Guaraní, thus expanding the reach of syntactic tools to underserved language communities.

%\bibliographystyle{acl_natbib}

%\bibliographystyle{acl_natbib}
%\bibliography{references}

% (rest of the document remains unchanged from previous version)

\section{Dataset and Experiments}
\label{sec:data_exp}
\begin{table}[t]
\scriptsize
\centering
\setlength{\tabcolsep}{4pt} % reduce column spacing
\begin{tabular}{llrr}
\toprule
\textbf{Datasets} & \textbf{Statistics} & \textbf{Spanish-English} & \textbf{Spanish-Guaraní} \\
\midrule
\multirow{2}{*}{Original} 
    & Sentences & $\approx 56{,}000$ & 1{,}140 \\
    & Tokens    & 242{,}475 & $\approx 17{,}100$ \\
   
\midrule
\multirow{2}{*}{Code-Switched} 
    & Sentences & 2{,}837 & 1{,}140 \\
    & Tokens    & 30{,}811 & $\approx 17{,}100$\\
\bottomrule
\end{tabular}
\caption{Sentence and token counts in the original and code-switched subsets of the Spanish-English and Spanish-Guaraní datasets.}
\label{tab:dataset_stats}
\end{table}

\subsection{Datasets}

We use two existing datasets for implementing our pipeline for BiLingua Parser and creating linguistically annotated CSW datasets. The first dataset we use is the Miami Corpus \citep{deuchar2014building}, a well-known Spanish-English dataset widely used in bilingualism and NLP research \citep{fricke2016primed, chi-bell-2024-pos-cs, soto2020codeswitching}. The second is the Spanish-Guaraní dataset from the shared task GUA-SPA: Guarani-Spanish Code-Switching Analysis \citep{chiruzzo2023overviewguaspaiberlef2023}, including social media and news content with spontaneous multilingual usage in a low-resource setting.

\paragraph{Miami Spanish-English Corpus.}
This dataset comprises transcribed spoken interactions between bilingual speakers. Each sentence is tokenized and annotated with a language tag, part-of-speech (POS) tag, and morphological features (e.g., \texttt{be.V.3S.PRES}). Metadata includes token index, sentence and utterance IDs, speaker identity, and filename. An example utterance with a switch into English is shown below:
\begin{quote}
\noindent \textbf{Speaker A:} \textit{la composición es increíblemente asociada a Joachim \underline{because} la tocó ahí primero.} \\
\hspace*{0.0em}[Eng – "The piece is strongly associated with Joachim \underline{because} he played it there first."]
\end{quote}

\paragraph{Spanish-Guaraní Dataset.}
This dataset contains Spanish-Guaraní utterances in social media and news contexts. Each example is a tokenized sentence, where every token is annotated with a language tag or named entity label (e.g., \texttt{gn} for Guaraní, \texttt{es-b-ul} for Spanish beginning token, \texttt{ne-b-org} for the beginning of an organization entity). An illustrative example from the dataset is shown below:
\begin{quote}
\noindent \textbf{@USER:} Movilización \uline{kakuaa opu'ãva} tiranía venezolana \uline{rehe}.\\
\hspace*{0.0em}[Eng – "A large mobilization \uline{rising up against} the Venezuelan tyranny."]
\end{quote}

\noindent
In this example, \texttt{@USER} is labeled as a named entity (\texttt{ne-b-per}), while tokens such as \textit{kakuaa opu'ãva} and \textit{rehe} are labeled as Guaraní and the rest as Spanish. The mixture of Guaraní and Spanish illustrates natural code-switching behavior.

\paragraph{Code-Switch Subset.}
To analyze syntactic behavior in mixed-language contexts, we automatically filtered for code-switched sentences in both of these datasets. A sentence was classified as code-switched if it contained at least two tokens from different language tags (e.g., one in English and one in Spanish). Table~\ref{tab:dataset_stats} summarizes the number of sentences and tokens in both the full and code-switched subsets for each dataset.

 %This distinction allows us to capture both single-token insertions and more structurally integrated alternations.

\subsection{Experimental Setup}

To generate syntactic annotations for these datasets, we developed a lightweight pipeline powered by GPT-4.1 (version \texttt{gpt-4.1-2025-04-14}). We use the OpenAI API with a deterministic configuration: \texttt{temperature=0}, \texttt{top\_p=1}, and \texttt{max\_tokens=3000}. Each prompt consists of a system instruction followed by a user message including the CSW sentence and a request for token-level annotation in UD format. This pipeline is detailed further in Section~\ref{sec:methd}. The Spanish-English dataset also includes conversational features typical of spontaneous speech, such as ellipsis, interjections, repetitions, and hesitations, which are known indicators of informal or spoken registers \citep{georgi-etal-2021-evaluating}. We flag such examples using a binary column \texttt{SPEC} to facilitate future syntactic and discourse-level studies that may benefit from separate treatment of these constructions.

Our resulting pipeline processes only the CSW subset of each dataset and outputs a CoNLL-like table with eight columns: token index (ID), token form (FORM), language tag (LANG), lemma (LEMMA), Universal POS tag (UPOS), syntactic head index (HEAD ID), syntactic head token (HEAD), and dependency relation (DEPREL). Native speakers of the respective language pairs reviewed and corrected the model outputs to ensure annotation accuracy. An example of this format, based on a code-switched sentence from the Spanish-English dataset, is shown in Table \ref{tab:cs_conllu_final}. The resulting datasets are released under a permissive open-source license to encourage further research in low-resource and multilingual parsing.

\begin{table}[h]
\centering
\resizebox{\columnwidth}{!}{
\begin{tabular}{lllllllll}
\toprule
\textbf{ID} & \textbf{Token Form} & \textbf{LANG} & \textbf{LEMMA} & \textbf{UPOS} & \textbf{HEAD ID} & \textbf{HEAD} & \textbf{DEPREL} \\
\midrule
1  & and     & en     & and     & CCONJ & 7  & same      & cc \\
2  & tú      & es     & tú      & PRON  & 3  & sabes     & nsubj \\
3  & sabes   & es     & saber   & VERB  & 7  & same      & conj \\
4  & it      & en     & it      & PRON  & 6  & was       & nsubj \\
5  & was     & en     & be      & AUX   & 7  & same      & cop \\
6  & not     & en     & not     & PART  & 5  & was       & advmod \\
7  & same    & en     & same    & ADJ   & 0  & root      & root \\
8  & .       & other  & .       & PUNCT & 7  & same      & punct \\
\bottomrule
\end{tabular}
}
\caption{UD-style annotation of the code-switched sentence “and tú sabes it wasn't the same,” (Eng. "and you know it wasn't the same").}
\label{tab:cs_conllu_final}
\end{table}

%\noindent The final annotated datasets, including raw and corrected outputs, will be made publicly available via a GitHub repository upon acceptance of this paper. 

\section{Methodology}
\label{sec:methd}

Our methodology integrates four components to build and analyze syntactically annotated code-switched data: (1) developing BiLingua Parser for generating UD annotations; (2) validating the annotations through expert review and evaluating accuracy; (3) conducting structural analysis on intra-sentential switch points; and (4) extending this framework to Low-Resource languages. Figure~\ref{fig:teaser} provides an overview of the full pipeline.

\subsection{Development of BiLingua Parser}
\label{sec:llm-tagging}
To create the BiLingua Parser, we used GPT-4.1 via the OpenAI API to generate UD annotations for CSW sentences. The process of generating accurate UD annotations is already a complex and time-consuming task for monolingual data, the challenge becomes even greater in the context of bilingual or CSW input. Therefore, the prompts for BiLingua Parser were carefully crafted using few-shot examples and refined through iterative testing, incorporating feedback from linguistic experts familiar with the targeted language pairs. Our prompts were specifically designed to handle the non-canonical structures typical of spoken and informal language, such as contractions, repetitions, incomplete sentences, and elliptical coordination. The model was instructed to produce token-level annotations based on the traditional CoNLL-U format that include ID, FORM, LANG, LEMMA, UPOS, HEAD ID, HEAD, and DEPREL. Full details of the prompt structures are provided in Appendix \ref{app:prompts} and \ref{app:prompt_spa_gua}.

\subsection{Handling of Informal Syntactic Structures}
In conversational and code-switched speech, non-canonical structures such as dropped words, hesitations, and merged tokens frequently occur \cite{georgi-etal-2021-evaluating}. These phenomena pose challenges for automatic dependency parsing, as many UD parsers assume well-formed, complete sentences. Here we describe how BiLingua Parser’s prompts account for these informal constructions so that resulting annotations remain linguistically coherent.  

\paragraph{Incomplete or Elliptical Sentences.}
Conversational speech between multiple speakers often consists of interruptions between dialogues leading to incomplete sentences or ellipses. We distinguish between truly incomplete sentences and elliptical ones that omit syntactic elements but remain interpretable. Table \ref{tab:incomplete} and \ref{tab:ellipses} show how we assign dependencies using \texttt{dep}, \texttt{orphan}, or \texttt{\_} in such cases.

\begin{table}[h]
\centering
\resizebox{\columnwidth}{!}{
\begin{tabular}{llllll}
\toprule
\textbf{FORM} & \textbf{LEMMA} & \textbf{UPOS} & \textbf{HEAD ID} & \textbf{HEAD} & \textbf{DEPEND} \\
\hline
It    & it     & PRON  & 2 & s'  & nsubj \\
's    & be     & AUX   & 0 & root   & root \\
the   & the    & DET   & 4 & end    & det \\
end   & end    & NOUN  & 2 & s'  & attr \\
of    & of     & ADP   & \_ & \_    & case \\
the   & the    & DET   & \_ & \_     & det \\
.     & .      & PUNCT & 2 & s' & punct \\
\bottomrule
\end{tabular}
}
\caption{UD tagging of an incomplete sentence [`It's the end of the...'] with missing final noun phrase.}
\label{tab:incomplete}
\end{table}

\begin{table}[h]
\centering
\resizebox{\columnwidth}{!}{
\begin{tabular}{llllll}
\toprule
\textbf{FORM} & \textbf{LEMMA} & \textbf{UPOS} & \textbf{HEAD ID} & \textbf{HEAD} & \textbf{DEPEND} \\
\hline
Me      & yo       & PRON  & 2 & gusta  & iobj \\
gusta   & gustar   & VERB  & 0 & root   & root \\
comer   & comer    & VERB  & 2 & gusta  & xcomp \\
y       & y        & CCONJ & 2 & gusta  & cc \\
a       & a        & ADP   & 6 & ella   & case \\
ella    & ella     & PRON  & 2 & gusta  & conj \\
bailar  & bailar   & VERB  & 6 & ella   & orphan \\
.       & .        & PUNCT & 2 & gusta  & punct \\
\bottomrule
\end{tabular}
}
\caption{UD tagging of an elliptical sentence [`Me gusta comer y a ella bailar' (Eng- `I like eating and she dancing.')] with gapping.}
\label{tab:ellipses}
\end{table}

\paragraph{Repetitions.}
In spoken interaction, repetitions often arise due to hesitation or self-correction. When repetitions occur, both instances are assigned the same syntactic role and head to preserve structural alignment. We use a similar approach in our prompting to handle repetitions in the dataset. See Table \ref{tab:repetition} for an example.

\begin{table}[h]
\centering
\resizebox{\columnwidth}{!}{
\begin{tabular}{llllll}
\toprule
\textbf{FORM} & \textbf{LEMMA} & \textbf{UPOS} & \textbf{HEAD ID} & \textbf{HEAD} & \textbf{DEPEND} \\
\hline
Yo      & yo     & PRON  & 4 & sé     & nsubj \\
yo      & yo     & PRON  & 4 & sé     & nsubj \\
no      & no     & PART  & 4 & sé     & advmod \\
sé      & saber  & VERB  & 0 & root   & root \\
.       & .      & PUNCT & 4 & sé     & punct \\
\bottomrule
\end{tabular}
}
\caption{UD tagging of a sentence  [`Yo yo no sé.' (Eng - `I I don't know.')] with hesitation and subject repetition.}
\label{tab:repetition}
\end{table}

\paragraph{Contractions and Punctuation.}
In traditional linguistic parsers, contractions (e.g., \textit{don't, they're}) are tagged by splitting them into their components. The prompt for BiLingua Parser instructed the LLM to follow the same approach for English tokens and assign proper dependency roles to each part. Additionally, punctuation was consistently attached to the root or main clause verb using the \texttt{punct} label. See Table \ref{tab:contract} for a typical output.

\begin{table}[h]
\centering
\resizebox{\columnwidth}{!}{
\begin{tabular}{llllll}
\toprule
\textbf{FORM} & \textbf{LEMMA} & \textbf{UPOS} & \textbf{HEAD ID} & \textbf{HEAD} & \textbf{DEPEND} \\
\hline
She     & she     & PRON  & 3 & go     & nsubj \\
did     & do      & AUX   & 3 & go     & aux \\
n't     & not     & PART  & 2 & did    & advmod \\
go      & go      & VERB  & 0 & root   & root \\
.       & .       & PUNCT & 3 & go     & punct \\
\bottomrule
\end{tabular}
}
\caption{UD tagging of a sentence [`She didn't go.'] illustrating contraction splitting.}
\label{tab:contract}
\end{table}

\subsection{Annotation Validation and Evaluation}
\label{sec:validation}
It is important to note that evaluating the BiLingua Parser-generated UD annotations on CSW Spanish-English and Spanish-Guaraní data is quite challenging due to the absence of established gold-standard datasets. We measured the annotation quality of our resultant datasets using the Labeled Attachment Score (LAS), which assesses both correct head assignment and dependency relation for each token, and we also report individual accuracy for UPOS and DEPREL tags. To compute these metrics, we compared model outputs against two reference sets:

\begin{enumerate}
  \item \textbf{Manually annotated gold standard.}  A small subset of sentences was selected at random and fully annotated by linguistic experts. Creating this bilingual gold standard is a tedious process, and it requires constructing complete parse trees and assigning UPOS, head indices, and dependency labels by hand. LAS was then calculated by comparing the LLM output to these expert annotations.
  
  \item \textbf{Human-revised LLM output.} In a faster second round, two bilingual annotators reviewed and corrected the model’s own parse outputs. Inter-annotator agreement on this subset reached Cohen’s Kappa of 0.85, indicating high consistency despite the structural ambiguity of CSW text. This approach accepts the LLM’s annotations if they fall within a linguistically plausible range, even when differing from canonical UD labels. 
\end{enumerate}

\noindent One of the reasons for adopting the second evaluation approach is that it provides a rigorous benchmark for assessing LLMs' performance on CSW contexts, particularly in the absence of pre-existing gold annotations. Another key motivation for this method is that the traditional method of LAS calculation for UD parsers does not account for semantic similarity between dependency labels or POS categories. For example, while the distinction between \texttt{AUX} and \texttt{VERB} is clearly defined in the UD guidelines (copulas and auxiliary verbs are to be tagged as \texttt{AUX} only), there are other cases where tagging ambiguity is more justified. Consider the verb \textit{"want"} in the sentence \textit{"I want to ride my bicycle"}. Depending on the analysis, \textit{"want"} may be treated as a main verb with a clausal complement (\texttt{ccomp}) or with an open clausal complement (\texttt{xcomp}), reflecting subtle differences in control and argument structure. Traditional LAS, however, penalizes such alternatives equally, even when both are linguistically reasonable. The expert review process accounts for such variation and tolerates plausible alternative annotations when they are linguistically motivated. To accommodate these subtleties, we treat sets of semantically related UD tags (see Table~\ref{tab:ud_similar_tags}) as equivalent. Differences within each group are not counted as errors under our human-aligned evaluation (see Appendix~\ref{sec:annotator_guidance} for annotator guidelines). As an additional baseline, we also trained a multilingual UD parser via sequence labeling (UDSL) to compare the results of BiLingua Parser; full experimental details are provided in Section \ref{app:udsl}.

\begin{table}[tbp]
\centering
\scriptsize
\renewcommand{\arraystretch}{1.2} % slightly tighter row spacing
\setlength{\tabcolsep}{4pt}       % reduce horizontal padding
\begin{tabularx}{\columnwidth}{@{}lX@{}}
\toprule
\textbf{Functional Domain} & \textbf{Semantically Similar UD Tags} \\
\midrule
Verbal Core & \texttt{root}, \texttt{aux}, \texttt{cop} \\
Clausal Complements & \texttt{xcomp}, \texttt{ccomp} \\
Discourse/Clause Linking & \texttt{parataxis}, \texttt{appos}, \texttt{conj}, \texttt{discourse}, \texttt{mark}, \texttt{advmod} \\
Adjectival/Clausal Modifiers & \texttt{amod}, \texttt{acl}, \texttt{acl:relcl} \\
Nominal Modifiers & \texttt{nmod}, \texttt{obl}, \texttt{advmod} \\
Numeric/Adjectival Modifiers & \texttt{nummod}, \texttt{amod} \\
Referential/Appositional Structures & \texttt{appos}, \texttt{nmod}, \texttt{conj} \\
\bottomrule
\end{tabularx}
\caption{Groups of semantically similar UD tags considered equivalent for evaluation purposes.}
\label{tab:ud_similar_tags}
\end{table}

\noindent Our experience with the evaluation process for BiLingua Parser's outputs suggests that the current UD evaluation metrics can be too rigid for complex, multilingual data such as dialogues or real-life conversational text. Developing a more flexible evaluation framework that systematically recognizes acceptable annotation variants would benefit future work on dependency parsing in code-switched and other non-standard text genres.

\subsection{Syntactic Analysis of Code-Switching}
\label{sec:cs-analysis}

To demonstrate the utility of our LLM-based annotated datasets for linguistic research, we conducted a structural analysis of intra-sentential code-switching, a phenomenon in which two languages are used within a single sentence or utterance \citep{poplack1980sometimes}, as it presents particularly interesting structural challenges for syntactic analysis. A switch point was defined as a token where the language tag differed from that of the preceding token. For each switch-in token, we extracted its part-of-speech (POS), dependency label, and language tag to study syntactic behavior at the boundary. 

We aggregated switch-in tokens and examined which syntactic roles (e.g., determiners, objects, discourse markers) are most commonly involved in switching. This analysis helps answer questions such as whether switches occur more often in determiner positions or whether object slots are more flexible across languages. It also enables us to draw structural generalizations about how different language pairs manage code-switching syntactically, particularly with respect to typologically distinct pairs like Spanish-Guarani, where strong differences in grammatical structure (e.g., head-marking, word order, affix richness) may affect switch behavior. Consider the following example:

\begin{center}
\textbf{\textit{I bought \uline{un} coche blanco.}}

[Eng-`I bought a white car.']
\end{center}

\noindent
Here, the switch-in token \textit{"un"} is labeled as a determiner (\texttt{det}), offering one instance of switching into a noun phrase. These cases are especially informative in Spanish-Guaraní, where mismatches such as article absence in Guaraní contrast with Spanish structures. To ensure a meaningful syntactic analysis, we filtered for code-switched sentences containing at least three tokens. This yielded 1,711 annotated Spanish-English sentences and 877 annotated Spanish-Guaraní sentences suitable for more informed analysis.

\subsection{Training a Universal Dependencies Parser with Sequence Labeling}
\label{app:udsl}

In addition to generating LLM-based annotation, we trained a multilingual dependency parser using a sequence labeling approach. It can be used as an alternate baseline for the task undertaken by the BiLingua Parser. We used the CoDeLin framework and fine-tuned \texttt{bert-base-multilingual-cased} with two encoding strategies: Relative (REL) and Absolute (ABS), following \citet{Roca2023ASF} to train this parser. The training data combined UD English EWT \citep{silveira14gold} and Spanish AnCora \citep{taule2008ancora} datasets. These were merged, shuffled, and split into training, development, and test sets. The data was then encoded into sequence labels using CoDeLin. The parser was trained for 30 epochs using a learning rate of $1\mathrm{e}{-5}$, batch size of 64, weight decay of 0.001, and Adam epsilon of $1\mathrm{e}{-7}$. We decoded the predictions into CoNLL-U format for evaluation using the standard CoNLL 2018 script \citep{zeman-etal-2018-conll}. This parser serves as a supervised benchmark for parsing performance in monolingual and CSW contexts.

\subsection{Extension of BiLingua Parser to Low-Resource Languages}
\label{sec:low-resource-extension}

We extended the BiLingua Parser to low-resource language pairs where no syntactically annotated code-switched data is available to train supervised parsers. The Spanish-Guaraní dataset serves as a case study. Motivated by the scalability of LLMs in low-resource settings, we used prompt-based UD annotation combined with native speaker review to bootstrap syntactic resources without requiring large annotated corpora. Prompt and architectural details are provided in Appendix~\ref{app:prompt_spa_gua}. This dataset presented unique challenges due to its length and complexity. Nonetheless, the parser generated meaningful annotations that enable valuable syntactic analysis for code-switching in under-resourced, typologically diverse languages.

%In this extension, we retained the original tokenization and sentence segmentation from the source Guaraní-Spanish dataset \cite{chiruzzo2023overviewguaspaiberlef2023}. We did not instruct the model to split morphologically complex tokens or restructure long sentences. As a result, some input sentences exceeded 50 tokens, significantly longer and more syntactically complex than the Spanish-English dataset, which averages 5 tokens per sentence. These long, clause-rich sentences posed a greater challenge for dependency parsing. Our goal in this initial phase was to assess how LLMs perform when applied directly to unprocessed, naturalistic code-switched data without additional preprocessing or segmentation. This provides a realistic test of the model’s ability to generalize to noisy, low-resource, and typologically diverse inputs.

\section{Results and Linguistic Analysis}
\label{sec:analysis}
\subsection{Results of BiLingua Parser}

Table~\ref{tab:las_comparison} compares the performance of BiLingua Parser on the labeled attachment score (LAS) metric on code-switched datasets. For Spanish-English, the LLM-based annotation achieved \textbf{76.32\%} score when compared with the Gold Annotation and \textbf{95.29\%} when compared with Human reviewed outputs, outperforming earlier models that reported LAS scores below 75\% (Sec. \ref{sec:related}). In addition to the Spanish-English evaluation, we also report results for the Spanish-Guaraní dataset, which represents one of the first attempts to syntactically annotate code-switched data involving this low-resource language. The Spanish-Guaraní dataset achieved LAS scores of \textbf{59.90\%} and \textbf{77.42\%}, respectively, on the two methods mentioned above. Notably, the Universal Dependencies Spanish-English model (UDSL), a general-purpose multilingual parser, achieved only \textbf{14.71\%} LAS when compared with the Gold Annotation, highlighting the limitations of off-the-shelf models when applied to code-switched data.

\begin{table}[h]
\centering
\footnotesize
\setlength{\tabcolsep}{4pt} % tighten column separation
\begin{tabular}{@{}lcc@{}}
\toprule
\textbf{Dataset} & \textbf{Gold Annotation} & \textbf{Human Review} \\
\textbf{} & \textbf{(LAS)} & \textbf{(LAS)} \\
\midrule
Spanish-English  & 76.32\%                           & 95.29\%                        \\
Spanish-Guaraní  & 59.90\%                           & 77.42\%                        \\
UDSL (Spa-Eng)  & 14.71\%                           & --\%                        \\
\bottomrule
\end{tabular}
\caption{Comparison of LAS before and after expert review on code‐switched data.}
\label{tab:las_comparison}
\end{table}

\begin{table}[h]
\centering
\footnotesize
\setlength{\tabcolsep}{4pt}
\begin{tabular}{@{}lccc@{}}
\toprule
\textbf{Dataset} & \textbf{UPOS} & \textbf{DEPREL} & \textbf{LAS} \\
\midrule
Spanish-English  & 99.54\%       & 97.14\%            &  95.29\%         \\
Spanish-Guaraní  & 84.21\%          & 59.90\%            & 59.90\%         \\
\bottomrule
\end{tabular}
\caption{UPOS, DEPREL, and overall LAS performance after expert revision.}
\label{tab:upos_deprel_las}
\end{table}

 We also carried out a detailed analysis of the human-reviewed output results to showcase the overall accuracy of the UD tags generated by the BiLingua Parser. As shown in Table~\ref{tab:upos_deprel_las}, the parser achieves high accuracy across UPOS, DEPREL, and LAS metrics (above \textbf{90\%}), particularly in the Spanish-English dataset. While the Spanish-Guaraní dataset shows slightly lower performance in dependency parsing, UPOS tagging remains strong, which is a great result for developing linguistic resources for low-resource languages. These results suggest that large language models can robustly handle syntactic analysis in bilingual contexts, outperforming both hybrid models and general-purpose multilingual parsers not specifically trained on code-switching.
\begin{NoHyper}
\begin{figure*}[t]
  \centering
  % --- UPOS plot ---
  \begin{subfigure}[t]{0.48\textwidth}
    \centering
    \includegraphics[width=\linewidth]{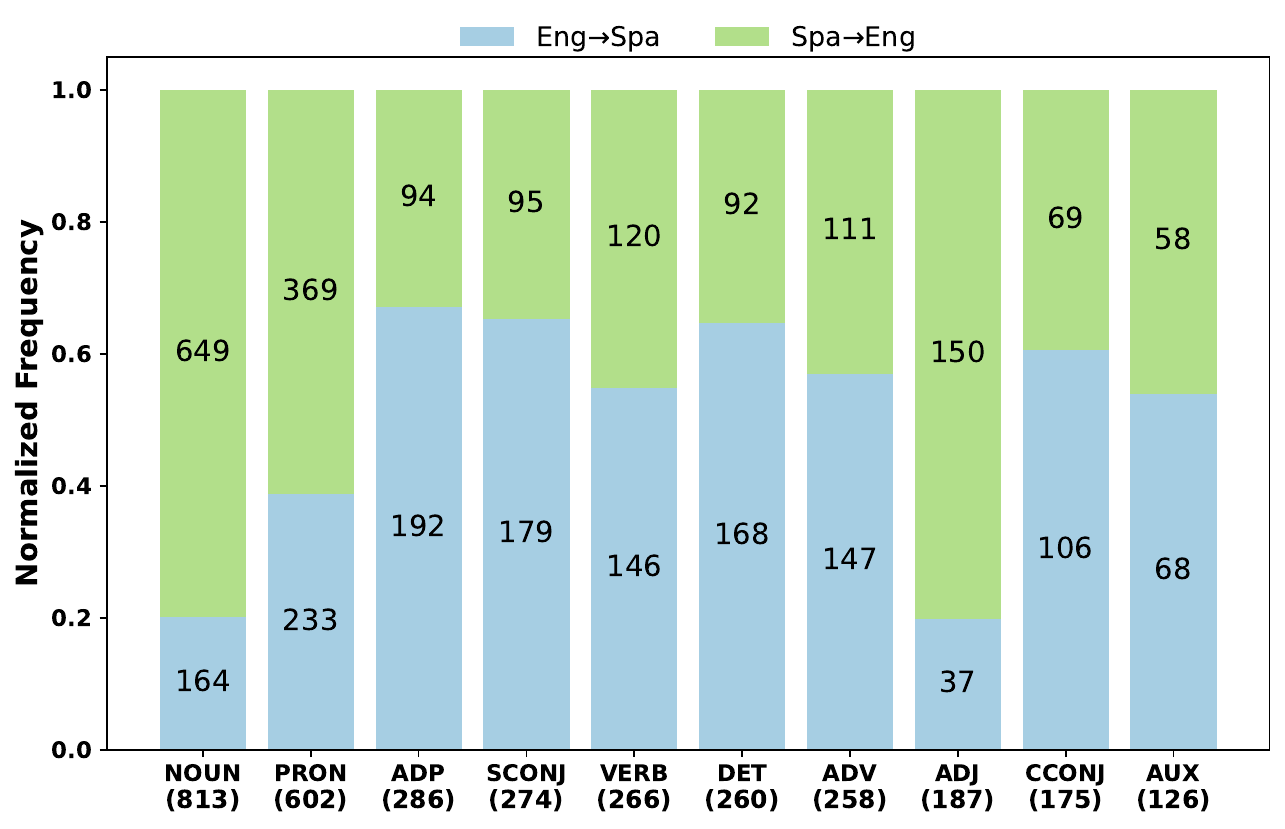}
    \caption{Normalized distribution of Universal Part-of-Speech (UPOS) tags at code-switch points in the Miami Spanish-English dataset. Notably, \texttt{NOUN}, \texttt{PRON}, and \texttt{ADP} are the most common categories at switch points, with \texttt{NOUN} exhibiting a high rate of switching from Spanish to English.}
    \captionsetup{hypcap=false}
    \label{fig:upos_switch}
  \end{subfigure}
  \hfill
  % --- DEPREL plot ---
  \begin{subfigure}[t]{0.48\textwidth}
    \centering
    \includegraphics[width=\linewidth]{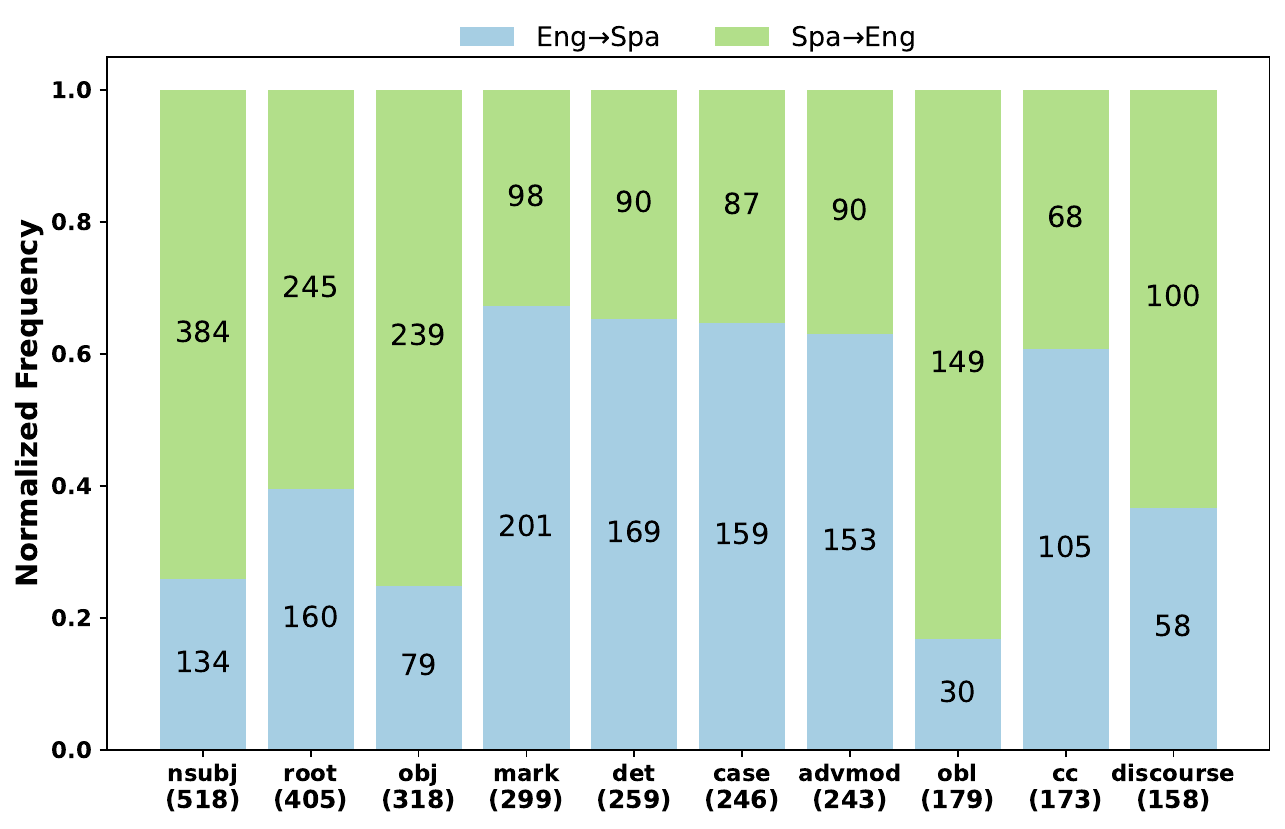}
    \caption{Normalized distribution of Universal Dependency Relations (DEPREL) at code-switch points in the Miami Spanish-English dataset. The most common relations at switch points are \texttt{nsubj}, \texttt{root}, and \texttt{obj}, indicating that syntactic subjects and core verbal arguments are key sites for switching.}
    \captionsetup{hypcap=false}
    \label{fig:deprel_switch}
  \end{subfigure}
  \caption{Normalized frequency distribution of syntactic categories (UPOS and DEPREL) at code-switch points across both switching directions. Bars show Eng$\rightarrow$Spa (blue) and Spa$\rightarrow$Eng (green) proportions. Absolute counts are shown inside bars; totals in parentheses.}
  \captionsetup{hypcap=false}
  \label{fig:upos_deprel_combined}
\end{figure*}
\end{NoHyper}
\subsection{Qualitative Error Analysis of LLM-based Dependency Parsing}

Although the prompt provides explicit guidelines for handling repetitions and ellipsis, LLM's responses remain inconsistent in applying these rules. It sometimes analyzes repetitions as coordinations, while in other cases it repeats the dependency structure for each clause. While both analyses are linguistically plausible, this inconsistency may affect the reliability of syntactic generalizations about code-switch points. We also observed inconsistencies in the analysis of functional verbs, including auxiliaries, modal verbs, and light verbs such as Spanish \textit{ser} (‘to be’). The LLM-based annotation oscillates between assigning these functional elements the syntactic role of \texttt{root} and attaching them to other verbal heads, indicating variability in the treatment of verbal dependency structures. 

Notably, native speaker feedback on the Guaraní data often identified morphologically complex words that should be split into multiple tokens for accurate syntactic and POS annotation. For instance, the token \textit{noñeguahëi} (‘not come’) was suggested to be split into a negational adverb and a verb, each with its own POS tag. This observation underscores the need for language-specific morphological preprocessing in low-resource and agglutinative languages and suggests that future improvements to LLM annotation pipelines may benefit from integrating morphological analyzers or token-splitting mechanisms tailored to these languages. Detailed examples investigating specific errors are provided in Appendix \ref{app:analysis}.

\subsection{Syntactic Generalizations at CSW Points in the English-Spanish dataset}

While prior research on the structural characteristics of code-switch points in NLP has largely focused on part-of-speech (POS) tags \cite{soto2020codeswitching}, our analysis advances this work by leveraging UD to capture syntactic roles at switch sites. Figures~\ref{fig:upos_switch} and~\ref{fig:deprel_switch} show the normalized distributions of UPOS and DEPREL tags across both switch directions. We find that subject positions (\texttt{nsubj}) are among the most frequent loci of switching, particularly in English-to-Spanish segments. Although this pattern is not consistently emphasized in the literature, it aligns with broader findings that permit switching at major syntactic boundaries, including clause-initial positions. Classic studies such as \cite{poplack1980sometimes, myersscotton2002contact} highlight noun phrases, especially determiners (\texttt{det}), modifiers (\texttt{amod}), and prepositions (\texttt{case}), as common switch sites when structural equivalence holds. Our results support this, showing frequent switches in the nominal domain and at clause boundaries (\texttt{mark}, \texttt{cc}, \texttt{discourse}). The prominence of \texttt{nsubj} may reflect language-pair-specific traits or discourse patterns, such as topic-prominence or left dislocation. These findings suggest that dependency relations uncover fine-grained switching patterns not captured by POS tags alone \cite{soto2020codeswitching}, and motivate the need for richer syntactic annotation in bilingual corpora.

Switching within the main verb or root predicate (i.e., the \texttt{root} in UD) has been considered highly constrained in the Spanish-English CSW literature. Early studies such as \citet{poplack1980sometimes} and models like the Matrix Language Frame \citep{myers1993duelling} argue that verb phrase boundaries are typically resistant to switching due to morphosyntactic incompatibilities between the two languages. Corpus-based studies \citep[e.g.,][]{toribio2001accessing, bullock2009trying, parafita2015subject} confirm that switching at or within the main verb is rare, with bilingual speakers favoring switches at clause boundaries. When switches do occur within the verbal domain, they tend to involve semantically transparent structures or frequent bilingual patterns. Our findings suggest that this restriction of the linguistic theory needs to be reexamined. %Switches that occur at nominal or adjectival main predicates are not rare in the Miami dataset. For instance, \textit{está demasiado \emph{empty}} (\texttt{root}): ‘It’s very/too empty’. 
The high frequency of code-switches at the root level may partly reflect parser errors, such as incorrectly analyzing modals or auxiliaries as roots. We acknowledge this limitation and plan to address it in future work by incorporating manual validation or model calibration strategies.

\subsection{Syntactic Generalizations at CSW Points in the Spanish-Guaraní dataset}
Our analysis of Spanish-Guaraní code-switching reveals broader syntactic flexibility than is typically observed in Spanish-English bilingualism. As shown in Figure \ref{fig:deprel-upos-emoji-vs-noemoji} (see Appendix \ref{app:spa_gua}), switch points in the Spanish-Guaraní data occur not only at canonical noun phrase boundaries, such as subjects (\texttt{nsubj}), objects (\texttt{obj}), and determiners (\texttt{det}), but also at clause-internal positions, including auxiliaries, modals, and root-level verbs. These sites are generally more resistant to switching in other language pairs. In contrast, the Spanish-English data (Figure~\ref{fig:upos_deprel_combined}) exhibit a more constrained switching pattern, largely centered on nominal boundaries and functional markers such as \texttt{mark} and \texttt{case}, with verbal heads showing lower susceptibility. The relative openness of Guaraní to verbal integration appears to license a wider range of switch locations. Further analysis, including a breakdown of emoji vs. non-emoji subsets and the role of discourse-level cues, is presented in Appendix~\ref{app:spa_gua}.

\section{Conclusion}
\label{sec:conc}

This work introduces, BiLingua Parser, a novel pipeline for syntactic annotation of code-switched data using LLMs, supported by expert human validation. By leveraging GPT-4.1 and linguistically informed prompting, we produced high-quality UD annotations for Spanish-English and Spanish-Guaraní code-switching. Our results show that LLM-based annotations outperform conventional parsers in syntactic accuracy, particularly at switch points where monolingual models typically fail. This performance gap is especially pronounced under our second evaluation method, which compares LLM outputs against human-revised annotations and does not penalize linguistically plausible variation. By incorporating groups of semantically similar dependency labels, this evaluation provides a more realistic benchmark for parsing in multilingual settings. Importantly, we release the first publicly available UD-annotated datasets for Spanish-English and Spanish-Guaraní CSW, addressing a critical gap in multilingual NLP resources. These datasets and our annotation methodology not only enable fine-grained analysis of code-switching behavior but also provide a foundation for advancing low-resource dependency parsing. %Future work will scale these efforts across additional language pairs and incorporate revisions to evaluation metrics that better reflect linguistic acceptability.

\section*{Limitations}

The UD framework provides a cross-linguistically consistent approach to syntactic annotation, but its complexity poses challenges for annotators unfamiliar with formal linguistic parsing conventions. Without such training, annotation quality may vary, and comparisons with other UD-based datasets may be less reliable. To ensure consistency and interoperability, we emphasize the importance of equipping native Guaraní speakers with detailed UD guidelines and hands-on annotation practice. This will support the creation of high-quality, linguistically grounded resources for low-resource languages. %that are compatible with the broader UD ecosystem.

\section*{Ethical Considerations}

Our work investigates the use of LLMs for syntactic annotation of code-switched language data, with a focus on Spanish-English and Spanish-Guaraní. While this research contributes to the development of more inclusive and multilingual NLP tools, it also raises several ethical considerations. The application of LLM-based syntactic annotation involves the risk of propagating model biases and structural inaccuracies, especially in under-resourced language contexts where gold-standard syntactic annotations are scarce. If such annotations are used for downstream tasks without human oversight, there is a danger of entrenching erroneous linguistic assumptions about bilingual speakers and their language practices. Naive or unsupervised deployment of LLMs in multilingual settings could unintentionally reinforce dominant-language structures or misrepresent code-switching norms. Before deploying such tools in real-world contexts, appropriate measures should be taken to ensure reliability and linguistic expertise. We have used AI assistants (Grammarly
and ChatGPT) to address the grammatical errors
and rephrase the sentences.

\section*{Acknowledgement}
We thank the anonymous annotators and reviewers for their constructive suggestions and help. We extend our gratitude to the Research Computing (RC) and Enterprise Technology at ASU for providing computing resources and access to the ChatGPT enterprise version for experiments. We acknowledge grants GAP (PID2022-139308OA-I00) funded by MICIU/AEI/10.13039/501100011033/ and ERDF, EU; LATCHING (PID2023-147129OB-C21) funded by MICIU/AEI/10.13039/501100011033 and ERDF, EU. CITIC, as a center accredited for excellence within the Galician University System and a member of the CIGUS Network, receives subsidies from the Department of Education, Science, Universities, and Vocational Training of the Xunta de Galicia. Additionally, it is co-financed by the EU through the FEDER Galicia 2021-27 operational program (Ref. ED431G 2023/01). 
%We thank Eliodora Verón, Instituto Superior de Lenguas, Universidad Nacional de Asunción for her help with the revision of the syntactic annotation of the Guaraní-Spanish dataset. 

\bibliography{custom}

\clearpage

\appendix
\section{Annotation Guidelines}
\label{sec:annotator_guidance}

\begin{table}[h]
\centering
\small
\renewcommand{\arraystretch}{1.2}
\begin{tabularx}{\linewidth}{@{}l l X@{}}
\toprule
\textbf{Tag} & \textbf{Label} & \textbf{Example(s)} \\
\midrule
\texttt{NOUN}  & Noun                    & \textit{house, tree} \\
\texttt{VERB}  & Verb                    & \textit{to run, to speak} \\
\texttt{ADJ}   & Adjective               & \textit{big, pretty} \\
\texttt{PRON}  & Pronoun                 & \textit{I, they} \\
\texttt{ADV}   & Adverb                  & \textit{quickly, well} \\
\texttt{ADP}   & Adposition              & \textit{in, under} \\
\texttt{DET}   & Determiner              & \textit{the, his/her} \\
\texttt{PROPN} & Proper noun             & \textit{Spain, Juan} \\
\texttt{NUM}   & Numeral                 & \textit{three, twenty} \\
\texttt{CCONJ} & Coordinating conjunction & \textit{and, but} \\
\texttt{SCONJ} & Subordinating conjunction & \textit{because, although} \\
\texttt{PART}  & Particle                & \textit{not, yes} \\
\texttt{INTJ}  & Interjection            & \textit{Hello!, Ugh!} \\
\texttt{PUNCT} & Punctuation             & \textit{., ?} \\
\texttt{other} & Miscellaneous           & \textit{Context-dependent} \\
\bottomrule
\end{tabularx}
\caption{Common UPOS tags provided during annotator training.}
\label{tab:agupos}
\end{table}
 
\begin{table}[h]
\centering
\small
\renewcommand{\arraystretch}{1.2}
\begin{tabularx}{\linewidth}{@{}l l X@{}}
\toprule
\textbf{Tag} & \textbf{Label} & \textbf{Example} \\
\midrule
\texttt{nsubj} & Nominal subject & \textit{She ran} → \textit{She} is the \texttt{nsubj} of \textit{ran} \\
\texttt{obj}   & Direct object    & \textit{I saw him} → \textit{him} is the \texttt{obj} of \textit{saw} \\
\texttt{iobj}  & Indirect object  & \textit{I gave her a book} → \textit{her} is the \texttt{iobj} \\
\texttt{root}  & Sentence root     & \textit{He left} → \textit{left} is the \texttt{root} \\
\texttt{det}   & Determiner        & \textit{The book} → \textit{The} is the \texttt{det} of \textit{book} \\
\texttt{case}  & Case marker       & \textit{in the house} → \textit{in} is the \texttt{case} of \textit{house} \\
\texttt{amod}  & Adjectival modifier & \textit{big house} → \textit{big} is the \texttt{amod} \\
\texttt{advmod} & Adverbial modifier & \textit{He ran quickly} → \textit{quickly} is the \texttt{advmod} \\
\texttt{conj}  & Conjunct in coordination & \textit{tea and coffee} → \textit{coffee} is the \texttt{conj} \\
\texttt{cc}    & Coordinating conjunction & \textit{tea and coffee} → \textit{and} is the \texttt{cc} \\
\bottomrule
\end{tabularx}
\caption{Key UD dependency relations introduced to annotators.}
\label{tab:agdeprel}
\end{table}

\noindent
We provided native speakers of Guaraní from Paraguay with a linguistic background with an overview of UD annotation scheme before annotation. This included explanations and examples for POS tags and DEPREL labels. A subset of the most relevant tags is listed in the Tables \ref{tab:agupos} and \ref{tab:agdeprel}. For Spanish-English annotations, native speakers of English and Spanish with a linguistic background were instructed to use the official Universal Dependencies documentation at \url{https://universaldependencies.org/} as a reference for POS and DEPREL labels during annotation.

\section{Prompts for the BiLingua Parser}
\begin{figure}[ht]
    \centering
    \includegraphics[width=0.91\linewidth]{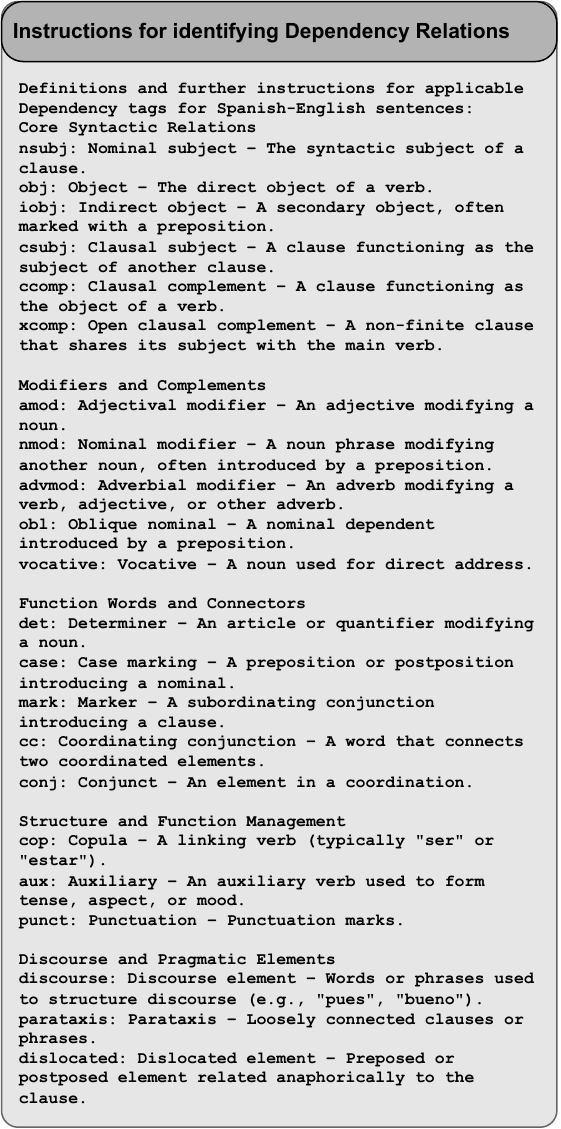}
    \caption{Dependency relation reference sheet provided to the model in the system prompt. The definitions follow UD conventions and include core \textbf{syntactic relations, modifiers, function words, clause-level structures, and discourse-related dependencies}. These definitions help constrain the model’s predictions to syntactically valid options for Spanish-English code-switch contexts.}
    \label{fig:inst}
\end{figure}

\begin{figure*}[ht]
    \centering
    \includegraphics[width=0.98\linewidth]{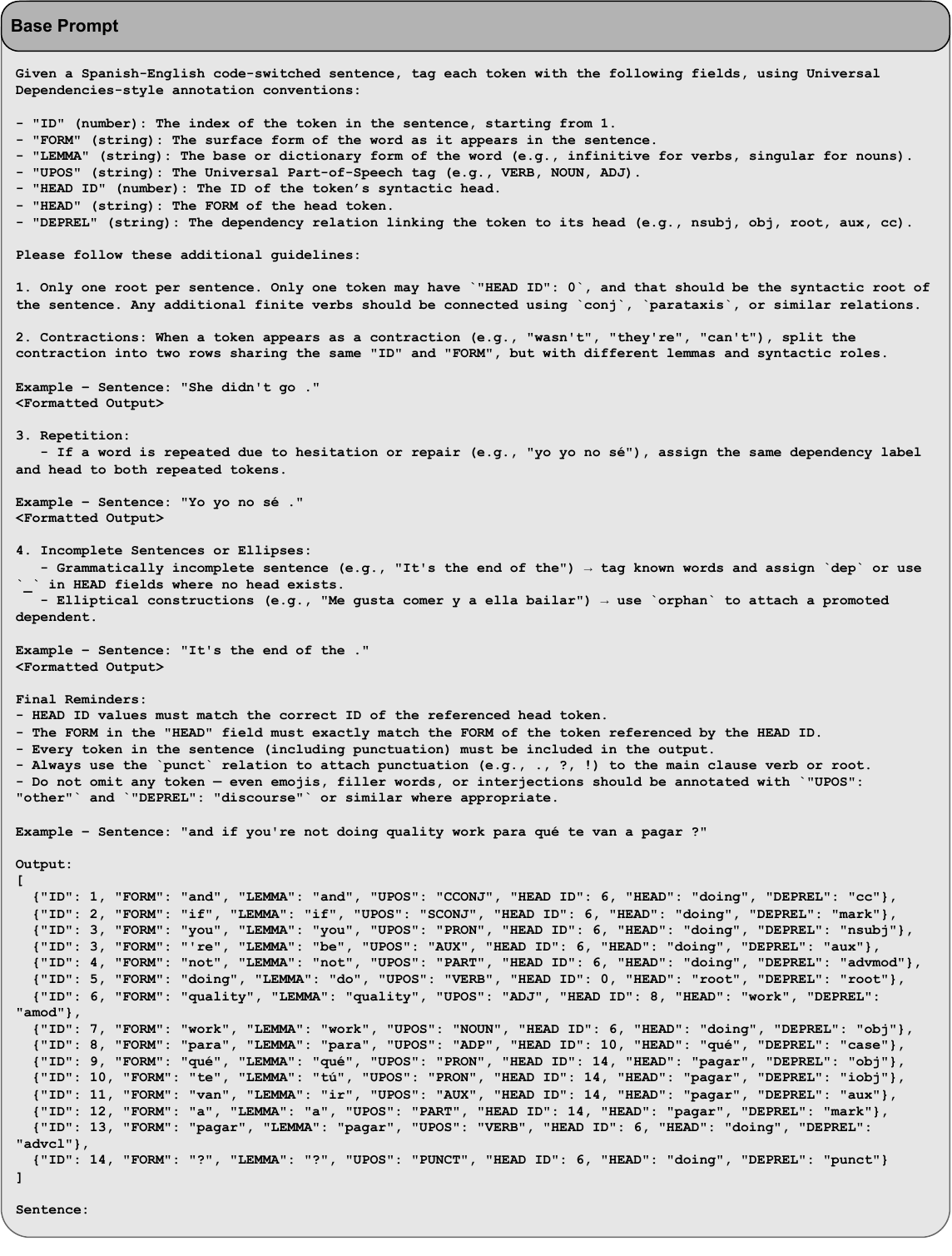}
    \caption{Prompt used to guide GPT in generating token-level UD annotations for \textbf{Spanish-English} code-switched sentences. The prompt outlines the required output format, including standard UD fields (\texttt{ID}, \texttt{FORM}, \texttt{LEMMA}, \texttt{UPOS}, \texttt{HEAD ID}, \texttt{HEAD}, \texttt{DEPREL}), and incorporates targeted instructions to address code-switching-specific phenomena. These include rules for handling English \textbf{contractions} (e.g., \textit{didn't} $\rightarrow$ \textit{did} + \textit{n't}), \textbf{disfluencies and repairs} (e.g., repeated tokens like \textit{yo yo}), \textbf{elliptical or incomplete} constructions, and \textbf{punctuation} attachment. The prompt ensures the sentence structure is valid by requiring \textbf{one root} token per sentence and giving rules for handling clausal and discourse-level dependencies. A fully formatted example illustrates the desired structure of GPT's response, aligning with UD conventions.}
    \label{fig:bprompt}
\end{figure*}

The figures below show the full prompt design used to guide the BiLingua parser. Figure \ref{fig:inst} presents a reference sheet of dependency relation definitions aligned with Universal Dependencies (UD) conventions. Figure \ref{fig:bprompt} shows the full base prompt given to the model, specifying the expected token-level format, as well as tailored instructions for handling special cases in code-switched input.
\label{app:prompts}

\clearpage

\begin{figure*}[ht]
    \centering
    \includegraphics[width=0.98\linewidth]{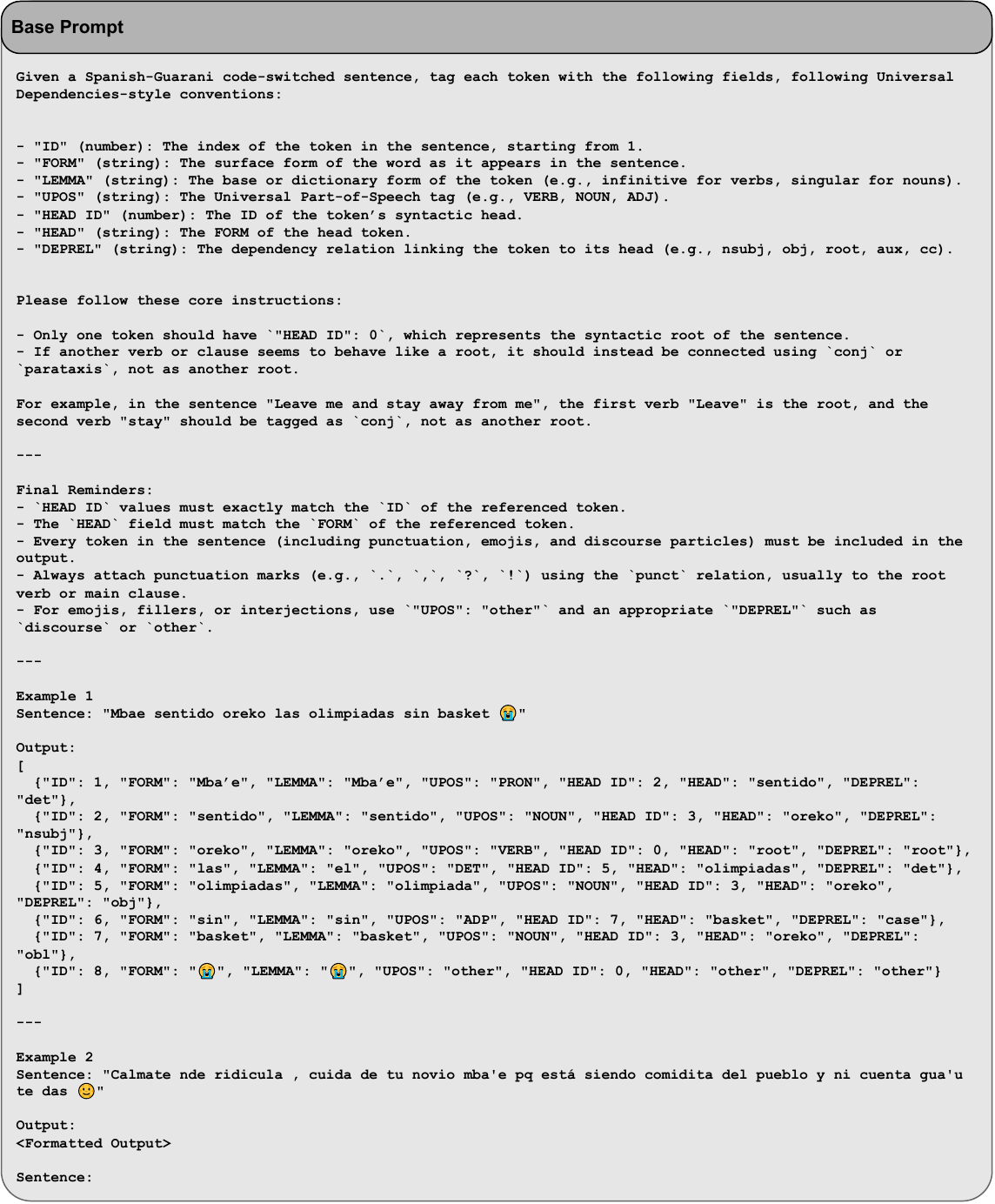}
    \caption{Prompt used to guide GPT in generating token-level UD annotations for \textbf{Spanish-Guaraní} code-switched sentences. The prompt defines the required Universal Dependencies (UD) output fields, \texttt{ID}, \texttt{FORM}, \texttt{LEMMA}, \texttt{UPOS}, \texttt{HEAD ID}, \texttt{HEAD}, and \texttt{DEPREL}, and enforces structural validity by requiring exactly \textbf{one syntactic root} per sentence. The instructions explicitly address how to attach additional verbs or clauses (e.g., using \texttt{conj} or \texttt{parataxis} rather than a second \texttt{root}) and how to treat punctuation and nonstandard tokens such as emojis or discourse particles using the \texttt{discourse} or \texttt{other} labels. Two fully formatted examples demonstrate how these conventions apply to mixed-language sentences, including Guaraní verbs and Spanish noun phrases. The prompt is designed to handle typologically diverse, low-resource input without preprocessing or morphological segmentation.}
    \label{fig:guarani-prompt}
\end{figure*}

\clearpage
\section{Architecture and Prompts for Spanish-Guaraní Dataset}
\label{app:prompt_spa_gua}

In constructing the Spanish-Guaraní UD annotations, we retained the original tokenization and sentence segmentation from the source dataset \cite{chiruzzo2023overviewguaspaiberlef2023}. The model was not instructed to split morphologically complex tokens or simplify the data. Consequently, many sentences exceeded 50 tokens and featured complex, clause-rich structures, in contrast to the shorter Spanish-English sentences which average around 5 tokens. This presented additional challenges for parsing accuracy. To address these challenges, we designed a task-specific prompt for Spanish-Guaraní code-switched input, shown in Figure~\ref{fig:guarani-prompt}. The prompt outlines the expected UD output format and includes targeted instructions for dependency structure validity, handling of discourse elements, and typologically diverse constructions. It enables the LLM to produce well-structured annotations without requiring preprocessing or morphological analysis, making it suitable for low-resource and morphologically rich language contexts.

\section{Extended Qualitative Analysis on CSW Results}
\label{app:analysis}

\noindent
Table~\ref{tab:ellip} shows an example of a syntactic structure containing both repetition and ellipsis. The phrase “they’re high enough so that él no se…” features repeated subject–copula constructions (“they’re”) across two overlapping clauses. The LLM inconsistently analyzes these repeated forms, sometimes attaching them in parallel, sometimes duplicating heads. It also treats the Spanish clause “él no se...” as an elliptical construction without resolving the final verb. This example illustrates the model’s challenges in managing discourse-level structures and maintaining syntactic coherence across long, code-switched utterances. Table~\ref{tab:repeat} illustrates another recurrent issue: inconsistent handling of repeated verbs. In the utterance “hay hay que dice o’clock somewhere,” the verb “hay” (‘there is’) appears twice, a common phenomenon in spontaneous speech. While both instances are valid, the LLM assigns the second instance a conjunct (\texttt{conj}) label instead of treating it as a disfluency or repetition of the root. This creates ambiguity in syntactic interpretation and points to the need for guidelines or preprocessing strategies for repeated tokens in code-switched input.

\begin{table}[h]
\centering
\resizebox{\columnwidth}{!}{%
\begin{tabular}{@{}c l l c l l@{}}
\toprule
\textbf{ID} & \textbf{FORM} & \textbf{LEMMA} & \textbf{HEAD} & \textbf{DEPREL} & \textbf{LANG} \\
\midrule
1  & but      & but      & 3  & cc      & eng \\
2  & I        & I        & 3  & nsubj   & eng \\
3  & think    & think    & 0  & root    & eng \\
4  & that     & that     & 7  & mark    & eng \\
\rowcolor{yellow!20}
5  & they’re  & they     & 7  & nsubj   & eng \\
\rowcolor{yellow!20}
5  & they’re  & be       & 7  & cop     & eng \\
\rowcolor{yellow!20}
6  & they’re  & they     & 7  & nsubj   & eng \\
\rowcolor{yellow!20}
6  & they’re  & be       & 7  & cop     & eng \\
7  & high     & high     & 3  & ccomp   & eng \\
8  & enough   & enough   & 7  & advmod  & eng \\
9  & so       & so       & 12 & mark    & eng \\
10 & that     & that     & 12 & mark    & eng \\
11 & él       & él       & 12 & nsubj   & spa \\
12 & no       & no       & 13 & advmod  & spa \\
\rowcolor{yellow!20}
13 & se       & se       & 7  & advcl   & spa \\
\rowcolor{yellow!20}
14 & .        & .        & 3  & punct   & --  \\
\bottomrule
\end{tabular}%
}
\caption{Dependency analysis of “But I think that they’re high enough so that él no se...” Highlighted rows show repeated subject–copula constructions and an elliptical adverbial clause.}
\label{tab:ellip}
\end{table}

\begin{table}[h]
\centering
\resizebox{\columnwidth}{!}{%
\begin{tabular}{@{}c l l l l l@{}}
\toprule
\textbf{ID} & \textbf{FORM} & \textbf{LEMMA} & \textbf{UPOS} & \textbf{HEAD} & \textbf{DEPREL} \\
\midrule
\rowcolor{yellow!20} 1  & hay       & haber    & VERB & 0  & root      \\
\rowcolor{yellow!20} 2  & hay       & haber    & VERB & 1  & conj      \\
5  & que       & que      & PRON & 2  & obj       \\
6  & dice      & decir    & VERB & 5  & acl:relcl \\
10 & o'clock   & o'clock  & NOUN & 6  & ccomp     \\
11 & somewhere & somewhere& ADV  & 10 & advmod    \\
\bottomrule
\end{tabular}%
}
\caption{Condensed UD analysis of “hay hay que dice o’clock somewhere.” Highlighted rows show the repeated verb “hay” handled inconsistently.}
\label{tab:repeat}
\end{table}

\section{Results for Emoji-Based Variation in Spanish-Guaraní dataset}
\label{app:spa_gua}
\begin{figure*}[t]
  \centering
  % Left: DEPREL (2 stacked subfigures)
  \begin{subfigure}[t]{0.48\textwidth}
    \centering
    \includegraphics[width=0.9\linewidth]{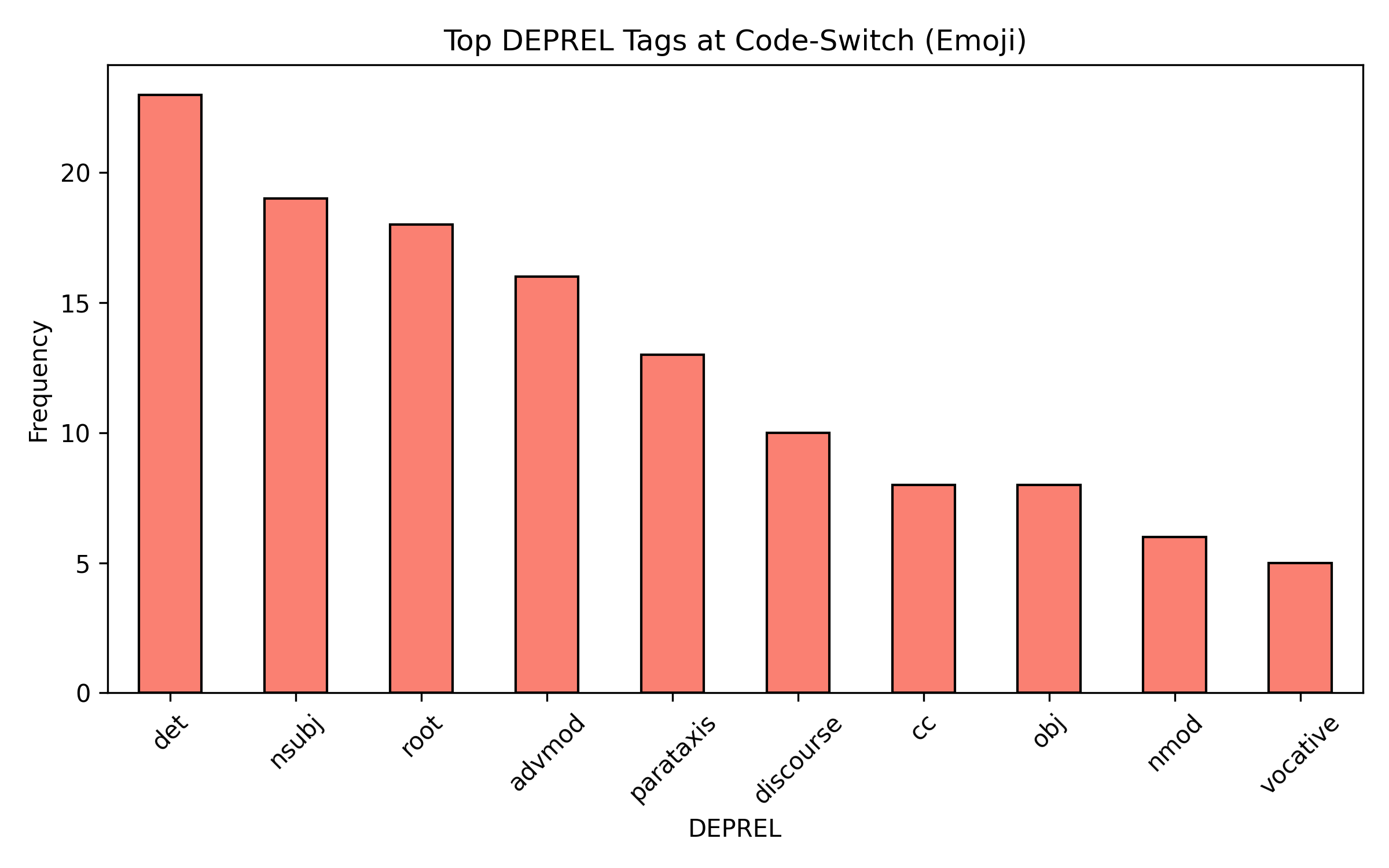}
    \caption{DEPREL + Emoji}
    \label{fig:deprel-emoji}
    \vspace{0.5em}
    \includegraphics[width=0.9\linewidth]{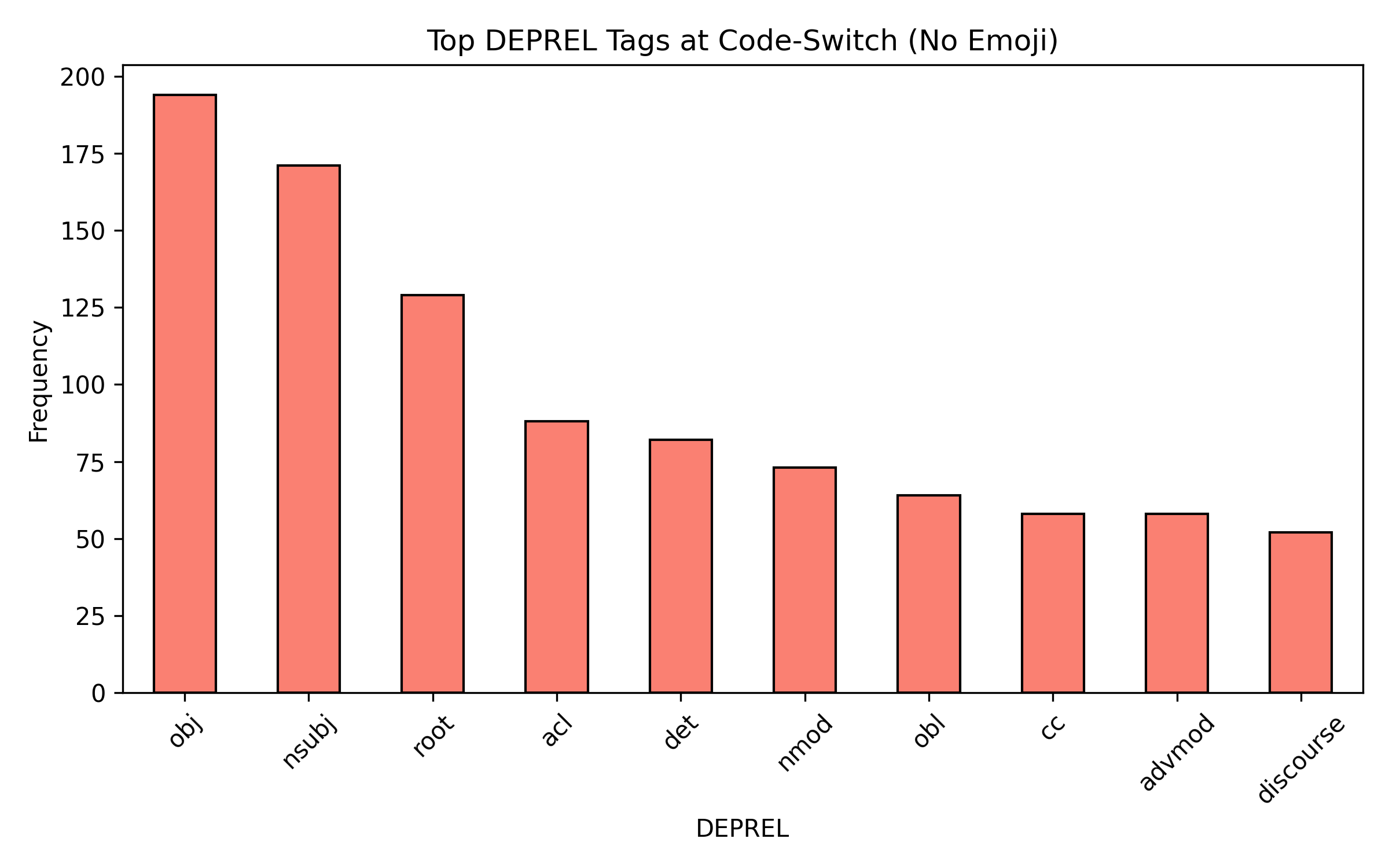}
    \caption{DEPREL – Emoji}
    \label{fig:deprel-noemoji}
  \end{subfigure}
  \hfill
  % Right: UPOS (2 stacked subfigures)
  \begin{subfigure}[t]{0.48\textwidth}
    \centering
    \includegraphics[width=0.9\linewidth]{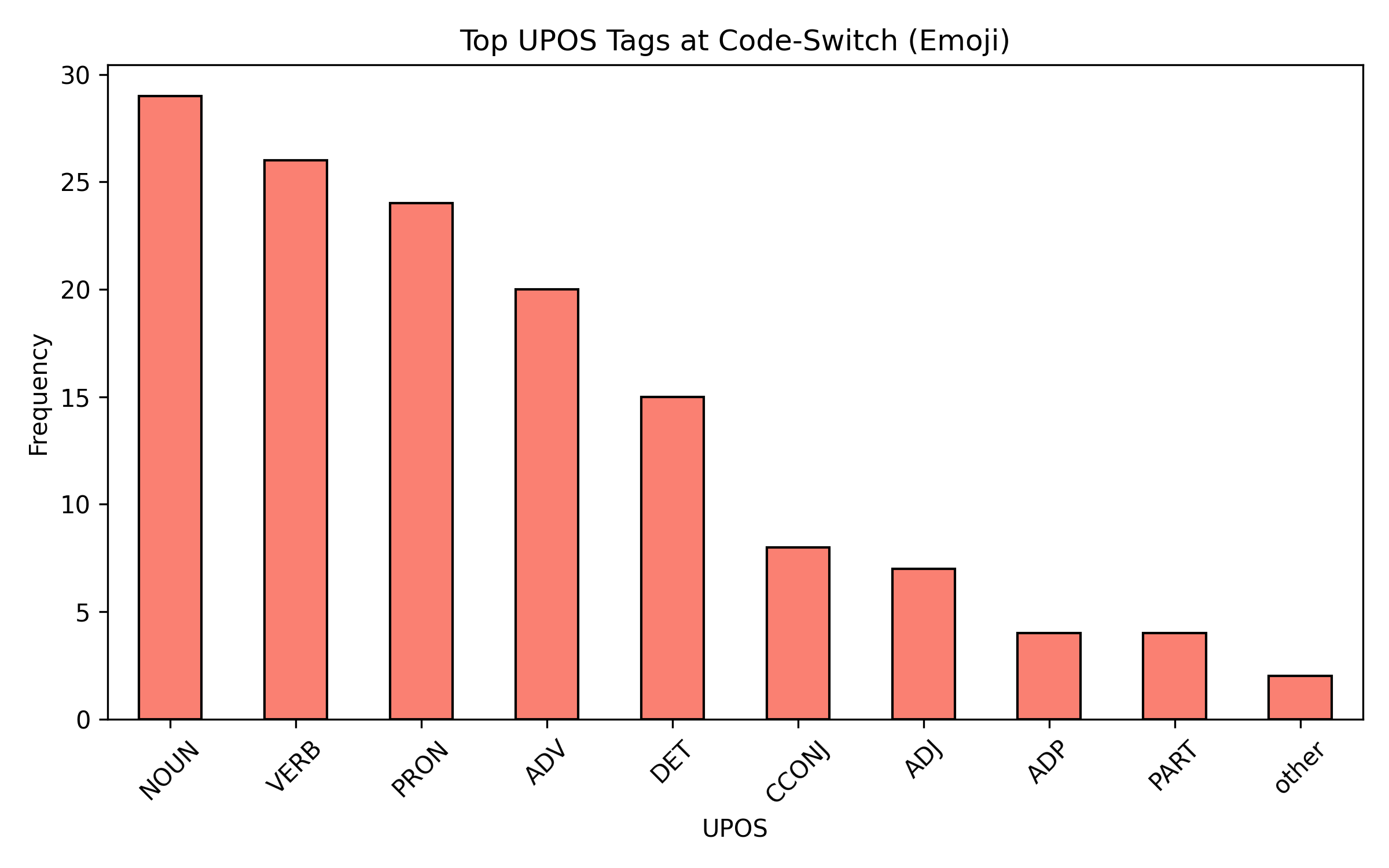}
    \caption{UPOS + Emoji}
    \label{fig:upos-emoji}
    \vspace{0.5em}
    \includegraphics[width=0.9\linewidth]{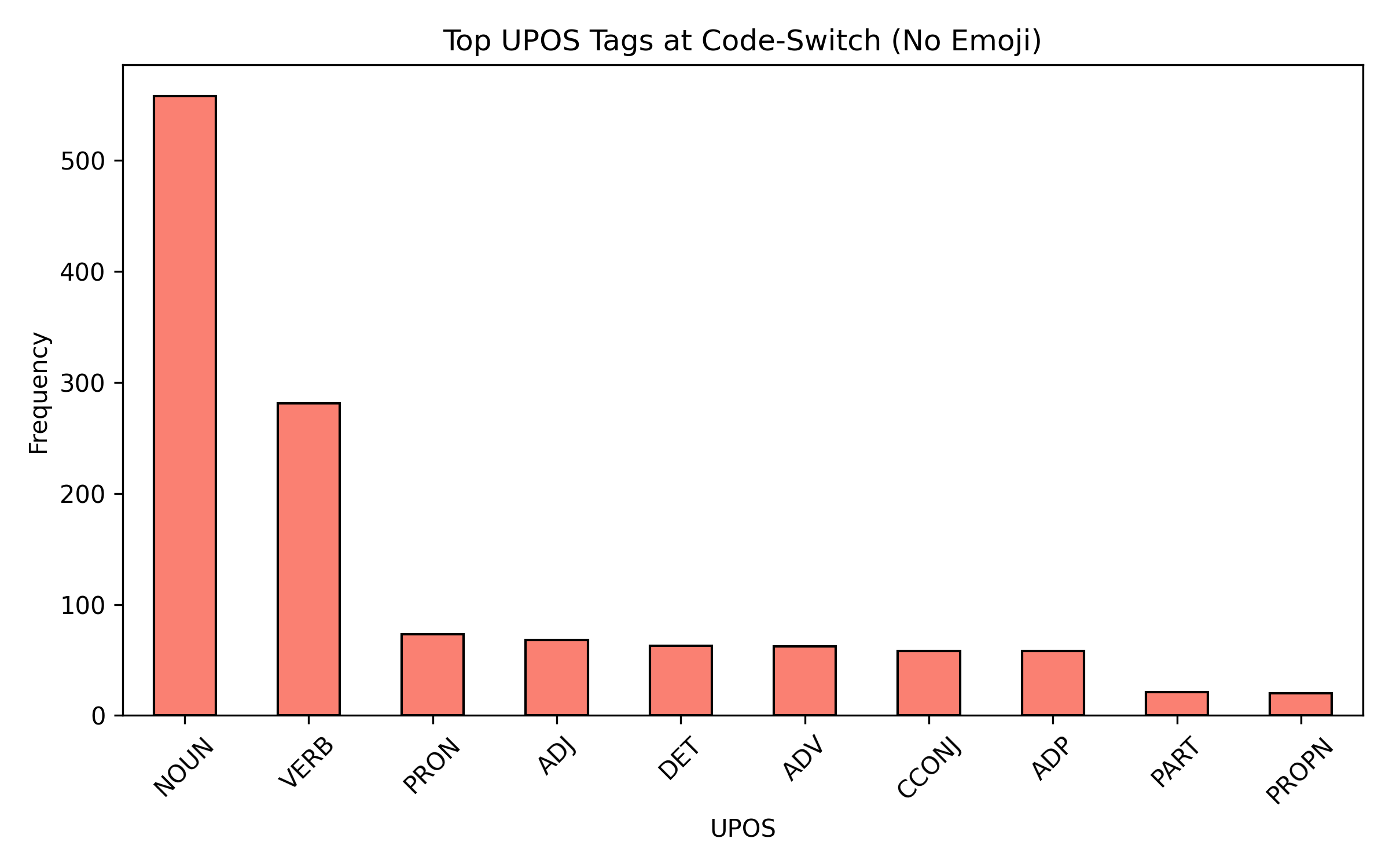}
    \caption{UPOS – Emoji}
    \label{fig:upos-noemoji}
  \end{subfigure}

  \caption{Distribution of DEPREL and UPOS tags at code-switch points in Spanish-Guaraní sentences, comparing emoji-containing and non-emoji subsets. (+ Emoji) refers to the subset of the dataset containing emojis, and (- Emoji) refers to the subset without the emojis.}
  \label{fig:deprel-upos-emoji-vs-noemoji}
\end{figure*}

To further understand discourse variation in Spanish-Guaraní code-switching, we divided the dataset into two subsets: messages with emojis and those without. This split approximates a difference in formality and expressiveness, with the emoji-containing subset representing more informal or emotionally expressive communication. Figure \ref{fig:deprel-upos-emoji-vs-noemoji} presents the top UPOS and DEPREL tags at code-switch points for both subsets. In the emoji-rich subset (Figures~\ref{fig:deprel-emoji}, \ref{fig:upos-emoji}), switching occurs frequently at discourse-sensitive syntactic roles such as \texttt{discourse}, \texttt{parataxis}, and stance-related verbs, in addition to traditional sites like \texttt{det}, \texttt{nsubj}, and \texttt{root}. This suggests that informal messages allow for more syntactic flexibility and that pragmatic context plays an important role in switch placement.

\noindent In contrast, the non-emoji subset (Figures~\ref{fig:deprel-noemoji}, \ref{fig:upos-noemoji}) reveals a more stable switching pattern, with concentration at canonical nominal positions such as \texttt{obj}, \texttt{nsubj}, \texttt{acl}, and \texttt{det}, and fewer instances of switching at clause-level discourse functions or verb heads. Together, these results support the observation that structural patterns of switching are not fixed but vary depending on the communicative context. Emoji usage appears to license greater fluidity in syntax, particularly at discourse-level transitions and pragmatically marked segments of bilingual speech.

\end{document}